\documentclass[pdflatex,sn-mathphys-num]{sn-jnl}
\usepackage{graphicx}%
\usepackage{multirow}%
\usepackage{amsmath,amssymb,amsfonts}%
\usepackage{amsthm}%
\usepackage{mathrsfs}%
\usepackage[title]{appendix}%
\usepackage{xcolor}%
\usepackage{textcomp}%
\usepackage{manyfoot}%
\usepackage{booktabs}%
\usepackage{algorithm}%
\usepackage{algorithmicx}%
\usepackage{algpseudocode}%
\usepackage{listings}%
\usepackage{amssymb}
\usepackage{amsmath}
\usepackage{times}
\usepackage{soul}
\usepackage{url}
\usepackage{booktabs}
\usepackage{algorithm}
\usepackage[switch]{lineno}
\usepackage{algpseudocode}%
\usepackage{amsfonts}
\usepackage{multirow}
\usepackage{listings}
\usepackage{xcolor}
\usepackage{booktabs}
\usepackage{array}
\usepackage{booktabs, multirow}
\theoremstyle{thmstyleone}%

\theoremstyle{thmstyletwo}%

\theoremstyle{thmstylethree}%

\begin{document}

\title[Article Title]{\textit{CoALFake}: Collaborative Active Learning with Human-LLM Co-Annotation for Cross-Domain Fake News Detection}

\author[1]{\fnm{Esma} \sur{Aïmeur}}

\author[2]{\fnm{Gilles} \sur{Brassard}}

\author*[3]{\fnm{Dorsaf} \sur{Sallami}}\email{dorsaf.sallami@umontreal.ca}

\affil{\orgdiv{Department of Computer Science and Operations Research }, \orgname{University of Montreal}, \orgaddress{\country{Canada}}}

\affil[1]{\href{https://scholar.google.com/citations?user=rBTFbxYAAAAJ&hl=fr}{Google Scholar Profile}}
\affil[2]{\href{https://scholar.google.com/citations?user=Rh7_srgAAAAJ&hl=fr&oi=ao}{Google Scholar Profile}}
\affil[3]{\href{https://scholar.google.com/citations?user=wUa3IWgAAAAJ&hl=fr}{Google Scholar Profile}}

\abstract{The proliferation of fake news across diverse domains highlights critical limitations in current detection systems, which often exhibit narrow domain specificity and poor generalization. Existing cross-domain approaches face two key challenges: (1) reliance on labelled data, which is frequently unavailable and resource-intensive to acquire and (2) information loss caused by rigid domain categorization or neglect of domain-specific features. To address these issues, we propose \textit{CoALFake}, a novel approach for cross-domain fake news detection that integrates Human-Large Language Model (LLM) co-annotation with domain-aware Active Learning (AL). Our method employs LLMs for scalable, low-cost annotation while maintaining human oversight to ensure label reliability. By integrating domain embedding techniques, the \textit{CoALFake} dynamically captures both domain-specific nuances and cross-domain patterns, enabling the training of a domain-agnostic model. Furthermore, a domain-aware sampling strategy optimizes sample acquisition by prioritizing diverse domain coverage. Experimental results across multiple datasets demonstrate that the proposed approach consistently outperforms various baselines. Our results emphasize that Human–LLM co‑annotation is a highly cost‑effective approach that delivers excellent performance. Evaluations across several datasets show that \textit{CoALFake} consistently outperforms a range of existing baselines, even with minimal human oversight.}


\maketitle

\section{Introduction}
The proliferation of fake news has become a global concern, with the rapid spread across social media platforms and influencing public opinion on critical issues such as elections, public health and climate change~\cite{sallami2023hype}. Therefore, in order to restrain the spread of false information, it is necessary to identify the falsity of false information before it spreads in a large number and take measures to stifle it in time~\cite{sallami2025exploring}.

Current fake news detection models are typically trained on domain-specific datasets, limiting their ability to adapt to new or underrepresented areas and causing significant performance drops with emerging events~\cite{li2023improving,aimeur2025too}. Although automated detection has advanced, challenges remain in cross-domain scenarios. Domain-specific models often fail to generalize due to variations in linguistic patterns, topics and terminology~\cite{silva2021embracing}. Additionally, domain shifts, arising from differences in vocabulary, emotions and style across domains, further restrict model adaptability~\cite{wang2023soft}.
\\ 
To detect fake news across diverse domains, early studies~\cite{li2023improving,lin2022detect} collected and manually labelled small datasets from emerging news domains. However, such approaches depend on labelled data from emerging domains, which is often unavailable and can be costly and time-consuming to obtain. Recent advancements in transfer learning and domain adaptation show potential, but they often require complex architectures and extensive fine-tuning~\cite{lu2021fantastically}, making them impractical for dynamic, real-world applications. Some studies, such as Ma \textit{et al.}~\cite{ma2024fake}, assign hard domain labels to news records, which can result in significant information loss. For instance, news records that span multiple domains cannot be adequately represented using rigid labelling schemes. Moreover, fake news detection models in real-world scenarios frequently encounter newly emerging, time-sensitive events for which no labelled data exists~\cite{wang2025fake}. 
\\ 
To address these limitations, this study proposes \textit{CoALFake}, a novel framework that combines active learning with Human-LLM co-annotation to enable scalable and effective cross-domain fake news detection. Our approach tackles the challenge of labelling news by leveraging LLMs to reduce annotation costs and human annotators to refine the most ambiguous or noisy records. Additionally, \textit{CoALFake} mitigates information loss by introducing a domain-agnostic classifier that balances domain-specific and cross-domain knowledge, ensuring robust generalization across diverse domains. Finally, through domain-aware sampling, we strategically select informative news records for annotation, ensuring the labelled dataset comprehensively represents a wide range of domains.

Our contribution can be summarized in the following ways: (1) A Human-LLM co-annotation framework that improves annotation efficiency and accuracy while fostering human-AI collaboration. (2) A domain-agnostic classifier that leverages multi-task learning to balance domain-specific and cross-domain knowledge. (3) A domain-aware Active Learning (AL) strategy that ensures balanced representation of all domains, addressing the challenge of cross-domain generalization. 
\section{Related Works}
\subsection{Coss-Domain Fake News Detection}
Numerous studies have explored cross-domain fake news detection. For example,~\cite{han2020graph} treat the detection of fake news across domains as a form of continual learning, where a model is progressively trained across multiple tasks. They analyse fake news through its spread patterns and employ continual learning strategies to tackle the challenges of cross-domain fake news detection. However, this methodology faces two primary constraints: (1) It presupposes the sequential arrival of news from different domains, which may not always align with real-world data flows and (2) it requires prior knowledge of the news domain, which often isn't available. Nan \textit{et al.}~\cite{nan2021mdfend} utilized expert evaluations across several domains to assess news items, employing a domain gate to synthesize these assessments into a final verdict. This method heavily relies on comprehensive human-annotated datasets. Building on this, Liang \textit{et al.}~\cite{liang2022fudfend} introduced the FuDFEND model, which assigns multi-domain labels to news items. The model extracts comprehensive feature vectors and employs a discriminator to ascertain the veracity of the news. Wang \textit{et al.}~\cite{wang2023soft} use soft domain labels for each news item but still restrict each item to a single domain. Similarly, Ma \textit{et al.}~\cite{ma2024fake} categorize news records using rigid, domain-specific labels.
The aforementioned methods assume that each news item is assigned to a single domain label. However, this assumption may not hold true given the complex semantics present in news articles and risks significant information loss by oversimplifying nuanced contextual details.
 
Recent studies~\cite{li2023improving,lin2022detect} focus on collecting and manually labelling small datasets from emerging news domains. These methods employ domain adaptation techniques to fine-tune trained models for emerging domains. However, they rely on the availability of labelled data, which can be costly, time-consuming and not always accessible. Rastogi \textit{et al.}~\cite{rastogi2021adaptive} propose an adaptive detection method that dynamically selects the optimal model for each domain. However, relying on a few handcrafted features may limit classifiers' ability to capture complex, evolving fake news patterns.

\subsection{Active Learning for Fake News Detection}
Although active learning is well-established in machine learning, its application to fake news detection remains limited. Most current systems rely on supervised learning~\cite{amri2021exmulf}, which requires large, costly and labour-intensive labelled datasets.

Some researchers apply traditional AL to fake news detection. For example, Wang \textit{et al.}~\cite{wang2020weak} use AL to efficiently select randomly high-quality training instances from a small initial labelled set. However, their method has two main limitations: reliance on a pre-trained model introduces bias and limited adaptability to evolving data distributions.
Alternative methods use hybrid architectures. Ren \textit{et al.}~\cite{ren2020adversarial} introduce a heterogeneous information network that combines content, author and topic for fake news detection. Their active learning framework leverages a graph neural network with hierarchical attention and an adversarial selector to identify high-value, diverse samples in each cycle. A related architecture by Barnab{\`o} \textit{et al.}~\cite{barnabo2023deep} focuses exclusively on social network propagation patterns while disregarding textual content entirely. Notably, Folino \textit{et al.}~\cite{folino2024towards} shifted focus to computational efficiency, demonstrating how AL can optimize memory and energy consumption in fake news detection systems. While existing AL methods for fake news detection remain confined to single-domain settings, Silva \textit{et al.}~\cite{silva2021embracing} represent the sole effort toward cross-domain detection. However, their approach incurs high annotation costs by relying extensively on human-labelled data. To address these limitations, we integrate LLMs as annotators to reduce manual labelling efforts. Prior works~\cite{xiao2023freeal,zhang2023llmaaa} employ LLMs in AL for automated annotation in other applications, but our framework uniquely combines human oversight with LLM-generated labels, ensuring scalability while mitigating risks of LLM hallucinations.

\section{Methodology}
\begin{figure}[h]
\centering
\includegraphics[width=\textwidth]{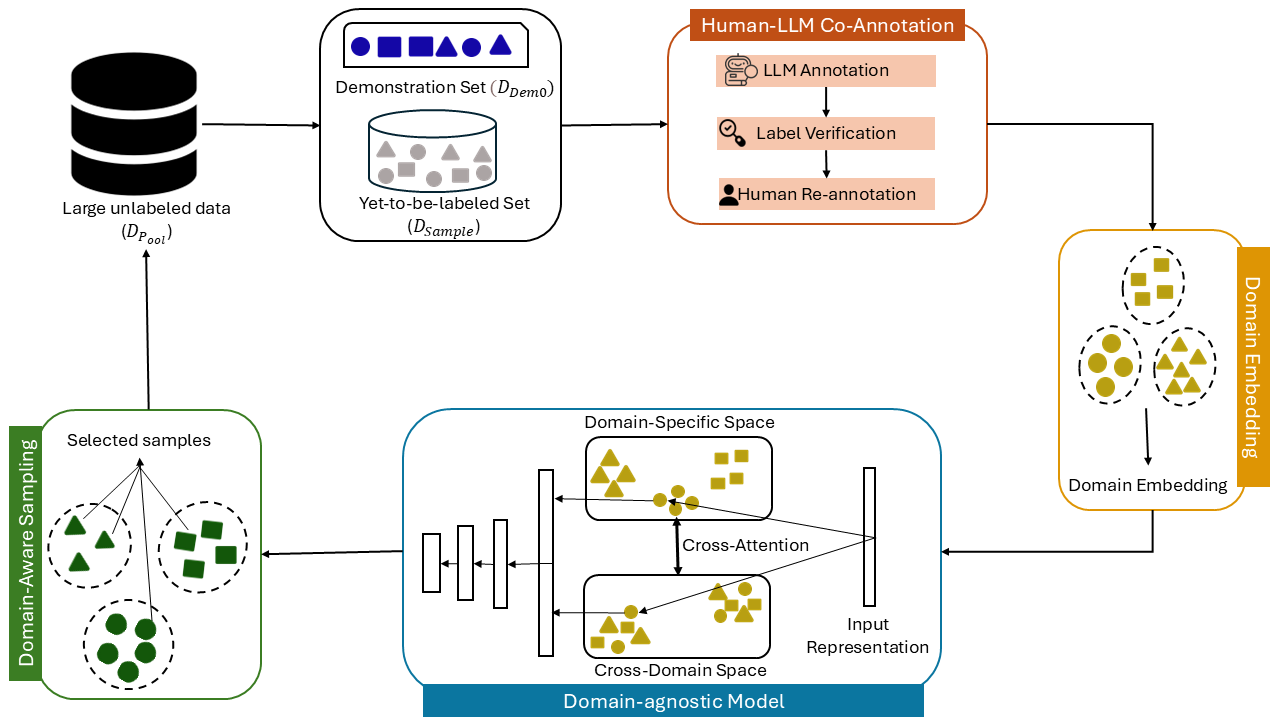}
\caption{Overview of \textit{CoALFake}. Data point shapes denote different domains, while colours indicate labelling phases: blue (demonstration set), gray (unlabelled), yellow (Human-LLM labelled) and green (post-classification labels). } 
\label{fig:method}
\end{figure}

\textit{CoALFake}, illustrated in Figure~\ref{fig:method}, comprises four core components designed to enhance cross-domain fake news detection. First, human-LLM co-annotation integrates human annotators and LLMs to generate high-quality labels collaboratively. Second, domain embedding maps news records into a domain-specific latent space, enabling the separation of domain-invariant and domain-specific features. Third, the domain-agnostic model leverages both domain-specific embeddings and shared semantic representations to improve generalization across domains. Finally, Active Domain-Aware Sampling strategically selects underrepresented domains to maximize coverage and enhance detection. A detailed overview of the full pipeline is provided in Algorithm~\ref{alg:pipeline}.

\begin{algorithm}[h]
\caption{\textit{CoALFake} Pipeline}
\label{alg:pipeline}
\begin{algorithmic}[1]
\State \textbf{Input:} Unlabelled Data ($D_{\text{Pool}}$), Demonstration Set ($D_{\text{demo}}$), LLM ($\mathcal{P}$), Human Annotator ($\mathcal{H}$)

\State \textbf{Output:} Trained Domain-Agnostic Classifier ($C$)

\State Initialize $D_{\text{labelled}} \gets \emptyset$, round $\gets 1$

\While{not convergent}
    \If{round $= 1$}
        \State Select $D_{\text{Sample}}$ using Eq.~\ref{eg:sample1} 
    \Else
        \State Select $D_{\text{Sample}}$ using Eq.~\ref{eg:sampleall} 
    \EndIf

    \State Retrieve $k$-NN examples $S = \{(x_i, y_i)\}_{i=1}^k$ from $D_{\text{demo}}$ using $k$-NN

    \State Perform in-context learning as Eq.~\ref{ed:LLM} to label $D_{\text{Sample}}$ 
    \State Perform label verification as Eq.~\ref{eq:labelverification}
    \State Send $D_{\text{Noisy}}$ to human annotators for re-annotation \
    \State Update $D_{\text{labelled}} \gets D_{\text{labelled}} \cup D_{\text{Sample}}$ 

    \State Compute domain embeddings as Eq.~\ref{eq:domainEmbed} 
    \State Train Domain-Agnostic Classifier $C$ using $D_{\text{labelled}}$ 
    \State Update $D_{\text{demo}}$ with new high-confidence samples 
    \State round $\gets$ round $+ 1$
\EndWhile

\State \textbf{Return:} Trained Classifier $C$
\end{algorithmic}
\end{algorithm}

Starting with unlabelled data $D_{\text{Pool}}$, the proposed system begins with the selection of representative samples  $D_{\text{Sample}}$. In the first round, samples are selected based on Eq.~\ref{eg:sample1}, while subsequent rounds utilize Eq.~\ref{eg:sampleall} for sample selection. The LLM annotates these samples through in-context learning as described in Eq.~\ref{ed:LLM}. To ensure label quality, label verification is performed according to Eq.~\ref{eq:labelverification} and any identified noisy labels are re-annotated by human experts. Domain embeddings are computed using Eq.~\ref{eq:domainEmbed} and the domain-agnostic model is iteratively trained with the refined dataset until convergence is achieved.

\subsection{Domain-Aware Sampling}\label{sec:cluster}
The model described above leverages both domain-specific and cross-domain knowledge to assess news truthfulness. However, its performance drops when detecting fake news in novel or infrequent domains, likely due to domain-specific vocabulary. To mitigate this, we propose an unsupervised active acquisition strategy that broadens domain coverage, ensuring robust performance across diverse domains.

Our methodology begins by clustering news articles to identify meaningful groups within the data. For each element \( x \) we first obtain an embedding representation \( \mathbf{z}_x\). Next, \( k \)-means clustering is performed on the set of embedded vectors \( \{\mathbf{z}_x\} \) to identify \( k \) distinct clusters. Each data point \( x \) is then assigned to a cluster \( \mathcal{C}_j \) based on its proximity to the nearest centroid. To assess the quality of the clustering, we use the Silhouette Score~\cite{shahapure2020cluster}.
Subsequently, the optimal number of clusters \( k \) is dynamically adjusted by iterating over a range \( [k_{\min}, k_{\max}] \), calculating Silhouette Scores for each value of \( k \) and selecting the \( k \) that maximizes the score, until the Silhouette Scores stabilize. 

Once clusters are established, we select representative samples for annotation. Instead of uniform sampling, we prioritize underrepresented domains by assigning inverse-proportional weights to each cluster based on its size:
\(
w_j = \frac{1}{|C_j| + \epsilon},
\)
where \( \epsilon \) prevents division by zero. These weights are normalized as:
\(
\hat{w}_j = \frac{w_j}{\sum_{m=1}^k w_m}.
\)

The number of samples from each cluster is determined by:
\[
m_j = \lceil M \cdot \hat{w}_j \rceil, \quad M = |D_{\text{Sample}}| \text{ (total samples)}.
\]

In traditional active learning methods, the initial training set is often randomly sampled, which can result in an unrepresentative subset. To address this cold-start problem, we propose a domain-aware, centroid-based initialization to ensure diverse and representative examples:
\begin{equation}\label{eg:sample1}
    D_{\text{Sample}} = \bigcup_{j=1}^{k} \text{Top-}m_j \big(x \in C_j, \text{ nearest to } \mu_j\big)
\end{equation}
After initializing the training set, subsequent iterations prioritize the most informative samples using an uncertainty-based approach. We select samples with the highest entropy, ensuring that the model focuses on instances where it is least confident:
\begin{equation}\label{eg:sampleall}
    D_{\text{Sample}} = \bigcup_{j=1}^{k} \text{Top-}m_j \big(x \in C_j, \text{ highest entropy } H(x)\big)
\end{equation}

For each sample \(x_i \in C_j\), the entropy is computed based on the classifier’s output probabilities as follows:

\[
H(x_i) = - \sum_{y=1}^{Y} p(y \mid x_i) \log p(y \mid x_i).
\]
We select \( m_j \) samples per cluster with the highest entropy (or least confidence \( 1 - \max_y p(y \mid x_i) \)).

\subsection{Human-LLM Co-Annotation}
\subsubsection{LLM Annotation}
In this step, we employ \textit{in-context learning} (i.e., \textit{prompting}) to enable the LLM to perform few-shot classification without the need for model fine-tuning. LLMs demonstrate excellent generalization capabilities and perform effectively as zero-shot and few-shot annotators \cite{brown2020language}. The process begins with a manually crafted prompt \( T(\cdot, \cdot) \), which is used in conjunction with demonstration examples \( S = \{(x_i, y_i)\}_{i=1}^{k} \) consisting of labelled examples and a query example \( x_q \) (i.e., an unlabelled news article). 
Based on this prompt, the LLM generates a prediction \( y_q \), which corresponds to the label that maximizes the conditional probability 
\begin{equation}\label{ed:LLM}
    y_q = \arg\max_y P_{\text{LM}}(y \mid T(S, x_q))
\end{equation}
Here, \( P_{\text{LM}} \) is the language model's conditional probability distribution over the possible labels, given the prompt. 

For the demonstration examples, we build upon the approach proposed by~\cite{liu2021makes}, which introduces a \( k \)-Nearest Neighbours (\( k \)-NN) retrieval strategy. This method first involves embedding the demonstration set \( D_{\text{Demo}} \) and the query sample into vector representations. Subsequently, it retrieves the nearest \( k \)  neighbours of the query to form its exemplars. The rationale behind this approach is that semantically similar examples may enhance the ability of LLMs to answer the query more accurately. We utilize Sentence-BERT \cite{reimers2019sentence} to construct these representations.
\subsubsection{Label Verification}
Once the labels are generated by the LLM, it is important to review each label and correct any inaccuracies. However, manually verifying every label produced would undermine the scalability and cost-efficiency benefits of using LLMs for annotation. Therefore, the next step is to select a subset of labels that are likely to be incorrect so they can be prioritized for human review.

The literature identifies two primary methods for assessing uncertainty in LLM outputs. One approach involves having the LLM self-assess or explicitly state its confidence level~\cite{lin2022teaching}, while the other method involves prompting the LLM multiple times with different formats to evaluate its consistency~\cite{wangself}. While these techniques are useful, they may not be as effective in our context due to (1) LLMs often displaying overconfidence~\cite{xiongcan} and (2) the predicted labels in our case being relatively consistent, as we focus solely on classification tasks.

Therefore, we adopt \textit{Confident Learning (CL)} to detect potentially erroneous labels. CL~\cite{northcutt2021confident} can estimate the joint distribution between the noisy labels (denoted by \( \tilde{y} \)) and the true latent labels (\( y^* \)) through the predicted probabilities from a model \( \theta \). 

Let \( X = \{(x, \tilde{y})\}^n \) be a noisy-labelled training set, with \( \hat{p}(\tilde{y}{=}j; x,\theta) \) as model-predicted probabilities. For class \( j \), compute threshold:
\[
t_j = \frac{1}{|X_{\tilde{y}{=}j}|} \sum_{x \in X_{\tilde{y}{=}j}} \hat{p}(\tilde{y}{=}j; x,\theta).
\]
Construct a confusion matrix \( C_{\tilde{y},y^*} \):
\[
C[i][j] = \left|\left\{ x \in X_{\tilde{y}{=}i} : \substack{\hat{p}(\tilde{y}{=}j; x,\theta) \geq t_j, \\ j = \arg\max_\ell \hat{p}(\tilde{y}{=}\ell; x,\theta) \geq t_\ell} \right\}\right|.
\]

Given the confident joint \( C_{\tilde{y},y^*} \),  we estimate \( Q_{\tilde{y},y^*} \) as:
\[
\hat{Q}[i][j] = \frac{\frac{C[i][j]}{\sum_{j'} C[i][j']} \cdot |X_{\tilde{y}{=}i}|}{\sum_{i',j'} \left( \frac{C[i'][j']}{\sum_{j''} C[i'][j'']} \cdot |X_{\tilde{y}{=}i'}| \right)}.
\]
For class \( i \), we identify potentially mislabelled samples by selecting: 
\begin{equation}\label{eq:labelverification}
    n \cdot \sum_{j \neq i} \hat{Q}[i][j]
\end{equation}  
samples with lowest \( \hat{p}(\tilde{y}{=}i; x \in X_i) \). These selected samples are flagged as the most likely to be wrongly labelled and they will be sent for human re-annotation.
\subsubsection{Human Re-annotation}
The LLM labels provide an initial reference, saving time for annotators. Human annotators then focus on reviewing and correcting a subset of samples flagged for potential label errors, identified during the label verification process.
\subsection{Domain Embedding}
Given a news record \( x \) where the domain label is unavailable, the proposed unsupervised domain embedding learning method leverages the textual data to represent the domain of \( x \) as a low-dimensional vector \( f_\text{domain}(x) \). 

Our methodology begins by clustering news articles (as in Section~\ref{sec:cluster}). Then, we compute the cosine similarity between \(z_x\) and each cluster centroid \(\mu_j\). The centroid \(\mu_j\) is the mean of the embedding vectors \(z_x\) for all \(x \in C_j\):
\(
\mu_j = \frac{1}{|C_j|} \sum_{x \in C_j} z_x
\).
The cosine similarity \(s(x, C_j)\) quantifies the alignment between \(z_x\) and \(\mu_j\):
\(
s(x, C_j) = \frac{z_x \cdot \mu_j}{\| z_x \| \, \| \mu_j \|}.
\)

After that, we convert similarities into probabilistic memberships $p(x \in C_j)$ using the softmax function, which ensures that the probabilities across all clusters sum to 1:
\[
p(x \in C_j) = \frac{\exp\left( s(x, C_j) \right)}{ \sum_{c=1}^{k} \exp\left( s(x, C_c) \right) }
\]
Here, $p(x \in C_j)$ reflects the semantic alignment of data point $x$ with cluster $C_j$, emphasizing semantic relationships rather than mere lexical overlap.
Finally, a domain embedding \( f_{\text{domain}}(x) \) for each \( x \) is computed by concatenating its likelihood of belonging to the clusters:
\begin{equation}\label{eq:domainEmbed}
    f_{\text{domain}}(x) = \left[ p(x \in c_1) \oplus p(x \in c_2) \oplus \ldots \oplus p(x \in c_k) \right]
\end{equation}
, where \( \oplus \) denotes concatenation.

Most prior works assign rigid labels to news records, leading to information loss, as some records span multiple domains. Hard labels fail to capture this nuance. In contrast, our approach uses low-dimensional embeddings to preserve domain complexities and retain valuable information.
\subsection{Domain-agnostic Model}
The news classification model begins by representing each news record $x$ as a feature vector $\mathbf{f}_{\text{input}}(x)$, derived from its textual content. To effectively capture both domain-specific and cross-domain knowledge, the model maps $\mathbf{f}_{\text{input}}(x)$ into two distinct subspaces. The first subspace, $\mathbf{f}_{\text{specific}}(x): \mathbf{f}_{\text{input}}(x) \rightarrow \mathbb{R}^d$, focuses on domain-specific information, while the second subspace, $\mathbf{f}_{\text{shared}}(x): \mathbf{f}_{\text{input}}(x) \rightarrow \mathbb{R}^d$, preserves cross-domain knowledge, where $d$ represents the dimension of the subspaces.

The concatenated representation $\mathbf{f}_{\text{concat}}(x) = \mathbf{f}_{\text{specific}}(x) \oplus \mathbf{f}_{\text{shared}}(x)$ is used to predict the news label $y_x$ through a decoder $\mathbf{g}_{\text{pred}}$, as follows: $ y'_x = \mathbf{g}_{\text{pred}}(\mathbf{f}_{\text{concat}}) $ and to reconstruct the input representation $\mathbf{f}_{\text{input}}(x)$ using a decoder $\mathbf{g}_{\text{recons}}, \mathbf{f'}_{\text{input}}(x) = \mathbf{g}_{\text{recons}}(\mathbf{f}_{\text{concat}}(x))$.

These operations are governed by two primary losses:
\begin{align*}
L_{\text{pred}} &= \text{BCE}(y_x, y'_x) \\
L_{\text{recon}} &= \|\mathbf{f}_{\text{input}}(x) - \mathbf{f'}_{\text{input}}(x)\|^2 \quad 
\end{align*}

To leverage domain-specific knowledge, we introduce a loss term \( L_{\text{specific}} \) that trains a decoder \( g_{\text{specific}} \) to reconstruct domain embeddings \( f_{\text{domain}}(x) \) from \( f_{\text{specific}}(x) \), ensuring \( f_{\text{specific}} \) captures domain-specific characteristics:  
\[
L_{\text{specific}} = \| f_{\text{domain}}(x) - g_{\text{specific}}(f_{\text{specific}}(x)) \|^2, 
\]
\[
\hat{g}_{\text{specific}}, \hat{f}_{\text{specific}} = \arg\min_{g_{\text{specific}}, f_{\text{specific}}} L_{\text{specific}}.
\]

In contrast, we train \(\mathbf{f}_{\text{shared}}\) to preserve cross-domain information by introducing a decoder \(\mathbf{g}_{\text{shared}}\) that predicts the domain of \(x\) from \(\mathbf{f}_{\text{shared}}(x)\). Meanwhile, \(\mathbf{f}_{\text{shared}}\) is optimized to mislead \(\mathbf{g}_{\text{shared}}\) by maximizing its prediction loss, ensuring a focus on cross-domain knowledge. This interaction follows a minimax game, defined as:  

\[
L_{\text{shared}} = \|\mathbf{g}_{\text{shared}}(\mathbf{f}_{\text{shared}}(x)) - \mathbf{f}_{\text{domain}}(x)\|^2
\]
\[
(\hat{\mathbf{g}}_{\text{shared}}, \hat{\mathbf{f}}_{\text{shared}}) = \arg\min_{\mathbf{f}_{\text{shared}}} \arg\max_{\mathbf{g}_{\text{shared}}} (-L_{\text{shared}})
\]

The model employs two cross-attention modules to enhance interaction between subspaces. Specific-to-Shared Attention enables \( f_{\text{shared}} \) to integrate domain-specific insights from \( f_{\text{specific}} \), while Shared-to-Specific Attention enriches \( f_{\text{specific}} \) with cross-domain knowledge. These mechanisms ensure effective information exchange.

The attention mechanism is defined as:
\[
\text{Attention}_{\text{s2s}} = \text{Softmax}\left(\frac{QK^\top}{\sqrt{d}}\right)V,
\]
where \( Q \), \( K \) and \( V \) are query, key and value matrices from the respective subspaces.
Attention outputs are merged via gated residual connections:
\[
f_{\text{shared}}' = f_{\text{shared}} + \gamma_{\text{s2s}} \cdot \text{LayerNorm}\left(W_{\text{out}} \cdot \text{Attention}_{\text{s2s}}\right),
\]
\[
f_{\text{specific}}' = f_{\text{specific}} + \gamma_{\text{s2s}} \cdot \text{LayerNorm}\left(W_{\text{out}}' \cdot \text{Attention}_{\text{s2s}}\right).
\]

Learnable gates \( \gamma_{\text{s2s}} \), activated via a sigmoid function.

To maintain subspace roles despite cross-attention, the model introduces two regularization terms. 

The Orthogonality Loss (\( L_{\text{ortho}} \)) preserves subspace distinction by minimizing cosine similarity:
\[
L_{\text{ortho}} = \| f_{\text{specific}}'^\top f_{\text{shared}}' \|^2.
\]

The Contrastive Loss (\( L_{\text{contrast}} \)) enforces dissimilarity by enhancing intra-class similarity within \( f_{\text{specific}} \) while reducing similarity with \( f_{\text{shared}} \):
\[
\mathcal{L}_{\text{contrast}} = -\log \frac{\exp \left( \text{sim} \left( f_{\text{specific}}', f_{\text{specific}}' \right) / \tau \right)}{\sum_{x \in \{ \text{specific}, \text{shared} \}} \exp \left( \text{sim} \left( f_{\text{specific}}', f_x' \right) / \tau \right)}.
\]

The final loss function \( \mathcal{L} \) integrates multiple components to balance their contributions:
\[
\mathcal{L} = \mathcal{L}_{\text{pred}} + \lambda_1 \mathcal{L}_{\text{recon}} + \lambda_2 \mathcal{L}_{\text{specific}} + \lambda_3 \mathcal{L}_{\text{shared}} + \lambda_4 \mathcal{L}_{\text{ortho}} + \lambda_5 \mathcal{L}_{\text{contrast}},
\]
where \( \lambda_1, \dots, \lambda_5 \) control the relative importance of each term.

To manage the minimax dynamics in \( \mathcal{L}_{\text{shared}} \), the final loss \( \mathcal{L}_{\text{final}} \) is optimized in two steps. First, all parameters except the domain classifier \( g_{\text{shared}} \) (\(\theta_1\)) are updated:
\[
\theta_{b1} = \arg\min_{\theta_1} \mathcal{L}_{\text{final}}(\theta_1, \theta_2).
\]
Then, the domain classifier parameters (\(\theta_2\)) are optimized to maximize \( \mathcal{L}_{\text{shared}} \), enhancing domain invariance:
\[
\theta_{b2} = \arg\max_{\theta_2} \mathcal{L}_{\text{final}}(\theta_{b1}, \theta_2).
\]

\section{Experiments and Results}
\subsection{Experimental Setup}
\subsubsection{Model Implementation}
Each news record \( x \) is represented as a feature vector \( \mathbf{f}_{\text{input}}(x) \), extracted using Sentence-BERT~\cite{reimers2019sentence}. This input is processed through two parallel subspaces: a domain-specific subspace \( \mathbf{f}_{\text{specific}}(x) \) and a shared subspace \( \mathbf{f}_{\text{shared}}(x) \), both implemented as single dense layers with ReLU activation. To enable interaction, a cross-attention mechanism with two multi-head attention modules (four heads each) facilitates information exchange. In specific-to-shared attention, the shared subspace acts as the query, while the domain-specific features serve as key and value, the roles are reversed for shared-to-specific attention. The attention outputs are merged with the original subspace features via gated residual connections, controlled by learnable sigmoid-activated parameters. The model components include \( \mathbf{g}_{\text{pred}} \), a two-layer feed-forward network with sigmoid activation in the final layer; \( \mathbf{g}_{\text{recons}} \), a two-layer feed-forward network with ReLU in the hidden layer and linear activation in the output layer and \( \mathbf{g}_{\text{shared}} \), a two-layer feed-forward network with sigmoid activation. The model is optimized using Adam with separate learning rates: \( 10^{-4} \) for the generator and \( 10^{-5} \) for domain classifiers\footnote{Code availability: The code is available on GitHub at \url{https://github.com/dorsafsallami/CoALFake}.}.
\subsubsection{Parameter Settings}
\begin{figure*}[ht]
\centering
\includegraphics[width=\textwidth]{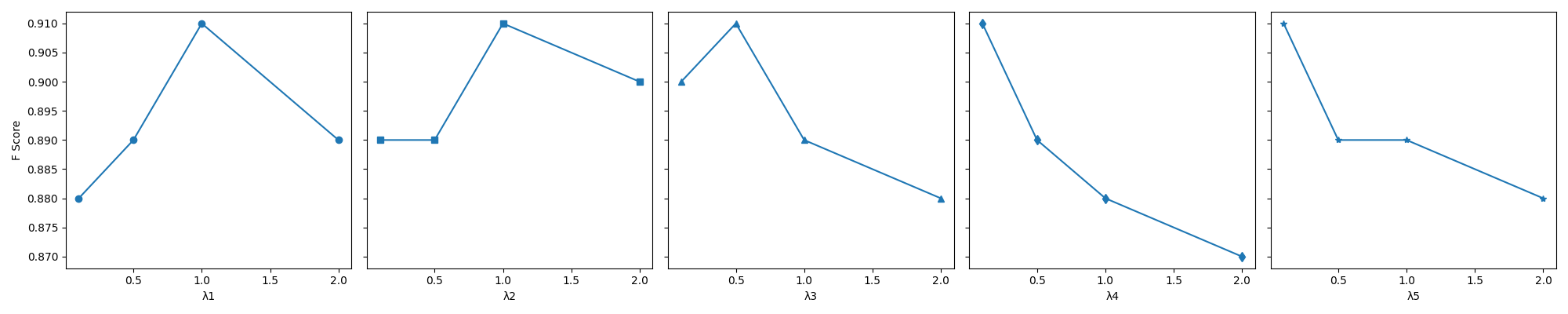}
\caption{Overall F1-Scores with different hyperparameters: \( \lambda_1, \lambda_2, \lambda_3, \lambda_4, \text{and} \lambda_5 \).  } 
\label{fig:lambda}
\end{figure*}

This section examines how modifying the model's hyperparameters influences its performance on fake news detection. Figure~\ref{fig:lambda} shows the model's behaviour for various values of \( \lambda_1, \lambda_2, \lambda_3, \lambda_4, \) and \( \lambda_5 \), each representing the weight assigned to a specific loss term. We observe that setting \(\lambda_1\) either too high or too low degrades performance, suggesting that the \(L_\text{recon}\) loss term should be given a moderate weight relative to the others. Similarly, performance declines when \(\lambda_2 < 1\) and when \(\lambda_3 > 1\). Additionally, increasing \(\lambda_4\) beyond 0.1 reduces the F1 score, and keeping \(\lambda_5\) small is essential to avoid over-clustering.

\begin{figure*}[ht]
\centering
\includegraphics[width=\textwidth]{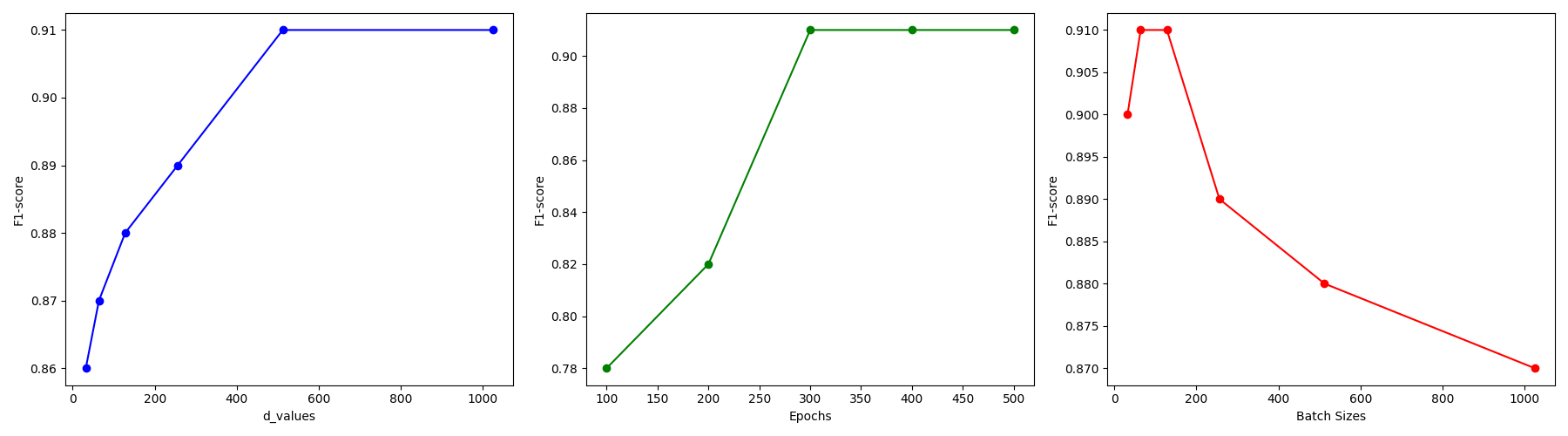}
\caption{Overall F1-Scores with different hyperparameters: Embedding Dimension (d), Epochs and Batch Size. } 
\label{fig:epochDbatch}
\end{figure*}

We examine the sensitivity of the model’s performance to other parameters: the latent dimension (\(d\)), the number of epochs and the batch size in Figure~\ref{fig:epochDbatch}. Overall, the model yields consistent performance for \(d > 512\), epochs \(> 300\) and batch size \(< 128\).

After performing a grid search, we set the hyperparameters in our model as follows: \( \lambda_1 = 1, \lambda_2 = 1, \lambda_3 = 0.5, \lambda_4 = 0.1, \lambda_5 = 0.1 \),  latent dimension \(d = 512\), epochs \(= 300\) and batch size \(= 128\). For baseline models, we used their default parameters from the original papers.

\subsubsection{Dataset}
We integrate three fake news datasets, PolitiFact (politics), GossipCop (celebrity) ~\cite{shu2020fakenewsnet} and CoAID (healthcare)~\cite{cui2020coaid}, to construct a comprehensive cross-domain news dataset. From each dataset, we randomly select 100 examples to form the demonstration set \( D_{\text{Demo}} \). The remaining data is split, with 75\% allocated to the training candidate pool \( D_{\text{Pool}} \) and 25\% reserved for testing. Performance is evaluated separately for each domain using its respective test instances, based on four key metrics: Accuracy (Acc), Precision (Prec), Recall (Rec) and F1 Score (F1).

\subsubsection{Model Baselines}
We compare our model with several widely used baselines:

\begin{itemize}
     \item \textbf{HPNF}~\cite{shu2019defend}: This model extracts multiple features, including structural and temporal characteristics, from a news article's propagation network to form its feature representation. A Logistic Regression classifier is then applied to distinguish between fake and real news. The \textbf{HPNF+LIWC} variant enhances this approach by integrating feature vectors from HPNF with those derived from LIWC.

    \item \textbf{AE}~\cite{silva2020embedding}: This method employs an Auto-Encoder architecture to learn latent representations of news records based on their propagation networks. These representations are subsequently used to identify fake news.

    \item \textbf{SAFE}~\cite{zhou2020similarity}: A multimodal method for fake news detection, SAFE learns separate latent representations for each modality of a news record while also constructing a joint representation that captures cross-modality insights.
    
    \item \textbf{EDDFN}~\cite{silva2021embracing}: This model leverages both domain-specific and cross-domain knowledge in news records to improve fake news detection across diverse domains. Additionally, an unsupervised technique selects a subset of unlabelled but informative news records for manual annotation.
    
    \item \textbf{MDFEND}~\cite{zhu2022memory}: This approach employs domain-specific experts to extract features from news articles, while domain gates assign varying weights to these experts based on their relevance. The final feature vector is obtained by aggregating the outputs of these experts.
    
    \item \textbf{FuDFEND}~\cite{liang2022fudfend}: This model begins by utilizing the final layer of the BERT Transformer block to transform news articles into word embedding vectors. A GRU module then generates multi-domain tags for each news item. The feature extraction module integrates these multi-domain tag features to construct the final comprehensive feature vector.

    \item \textbf{DITFEND}~\cite{nan2022improving}: This model transfers coarse-grained domain-level knowledge by training a general model on data from all domains using a meta-learning approach. To facilitate fine-grained instance-level knowledge transfer, a language model is trained specifically on the target domain. This model evaluates the transferability of each data instance from the source domains and re-weights their contributions accordingly.
    \item \textbf{SLFEND}~\cite{wang2023soft}: This model enhances feature extraction through the use of soft labels. A novel Leap GRU mechanism filters out irrelevant words, allowing the membership function module to generate soft labels for each news item. These soft labels aid in extracting multi-domain features, leading to a comprehensive feature representation.  
\end{itemize}

To ensure a fair comparison, all baseline models were trained using the same mixed-domain dataset (PolitiFact, GossipCop, and CoAID) as \textit{CoALFake}. This consistent training setup eliminates domain-related variance and ensures that performance differences reflect model design rather than differences in data exposure.
\subsubsection{Sampling Strategy Baselines}\label{sampling}
We compare our proposed sampling strategy with some common strategies for thorough comparisons: 
\begin{itemize}
    \item \textbf{Random Selection}; 
We use random selection as a baseline, which samples uniformly from \( D_{\text{Pool}} \). Since the pool data and test data generally share the same distribution, the sampled batch is expected to be i.i.d.~(independent and identically distributed) with the test data.

\item  \textbf{Maximum Entropy}:
Entropy is a widely used measure of uncertainty~\cite{settles2009active}. Data points with the highest entropy according to the model \( M \) are selected for annotation. The selection is based on the following criterion:
\[
\arg \max_{x \in D_{\text{Pool}}} \left( - \sum_{y \in Y} P_M(y|x) \log P_M(y|x) \right).
\]

\item \textbf{Least Confidence}: Culotta \textit{et al.}~\cite{culotta2005reducing} propose a method where examples are ranked based on the probability assigned by \( M \) to the predicted class \( \hat{y} \). The data point with the highest confidence (i.e., the model's least uncertainty) is chosen for annotation. The selection is made according to:
\[
\arg \max_{x \in D_{\text{Pool}}} \left( 1 - P_M(\hat{y}|x) \right).
\]

\item  \textbf{K-Means Diversity Sampling}: 
Diversity sampling aims to select batches of data that are heterogeneous in the feature space. Following~\cite{yuan2020cold}, we apply $k$-means clustering to the L2-normalized embeddings of \( M_4 \), and then sample the nearest neighbours of the \( k \) cluster centres.
\end{itemize}

\subsubsection{Co-annotation Setup}
We adopt OpenAI's GPT-3.5-Turbo language model as the backbone for the LLM. For the demonstration examples, we apply $k$-NN example retrieval, setting $k$ to 5, consistent with prior experiments~\cite{liu2021makes}. For the label verification model, we use a multinomial logistic regression classifier to identify label errors~\cite{northcutt2021confident}. The built-in stochastic gradient descent optimizer in the open-sourced fastText library~\cite{joulin2017bag} is used with the following settings: initial learning rate = 0.1, embedding dimension = 100 and $n$-gram = 3. Out-of-sample predicted probabilities are obtained via 5-fold cross-validation. For the re-annotation, we select 20\% of noisy labels to be re-annotated by humans after testing different values (Section~\ref{sec:annotation}).
\subsection{Results}
\textbf{Overall Results:}
 The results, as shown in Table~\ref{tab:results}, demonstrate that \textit{CoALFake} consistently outperforms all other models in terms of accuracy, precision, recall and F1 score. 

\begin{table}[h]
\centering
\caption{Evaluation results of \textit{CoALFake} and baseline models on three datasets.}\label{tab:results}
\footnotesize
\begin{tabular}{lllllllllllll}
\toprule
\multicolumn{1}{c}{\multirow{2}{*}{Model}}        & \multicolumn{4}{c}{Politifact}                                                                        & \multicolumn{4}{c}{Gossipcop}                                                                         & \multicolumn{4}{c}{CoAID}                                                                             \\ \cmidrule{2-13} 
\multicolumn{1}{c}{}                              & \multicolumn{1}{c}{Acc} & \multicolumn{1}{c}{Prec} & \multicolumn{1}{c}{Rec} & \multicolumn{1}{c}{F1} & \multicolumn{1}{c}{Acc} & \multicolumn{1}{c}{Prec} & \multicolumn{1}{c}{Rec} & \multicolumn{1}{c}{F1} & \multicolumn{1}{c}{Acc} & \multicolumn{1}{c}{Prec} & \multicolumn{1}{c}{Rec} & \multicolumn{1}{c}{F1} \\ \midrule
HPNF \cite{shu2019defend}        & 0.69                    & 0.69                     & 0.68                    & 0.68                   & 0.72                    & 0.70                     & 0.68                    & 0.69                   & 0.90                    & 0.65                     & 0.69                    & 0.67                   \\ 
HPNF + LIWC \cite{shu2019defend} & 0.70                    & 0.72                     & 0.70                    & 0.71                   & 0.73                    & 0.71                     & 0.70                    & 0.70                   & 0.91                    & 0.68                     & 0.70                    & 0.69                   \\ 
AE \cite{silva2020embedding}     & 0.78                    & 0.78                     & 0.77                    & 0.77                   & 0.83                    & 0.82                     & 0.80                    & 0.81                   & 0.92                    & 0.68                     & 0.67                    & 0.67                   \\ 

SAFE \cite{zhou2020similarity}   & 0.79                    & 0.78                     & 0.77                    & 0.77                   & 0.83                    & 0.82                     & 0.79                    & 0.80                   & 0.93                    & 0.75                     & 0.74                    & 0.74                   \\ 
EDDFN \cite{silva2021embracing}  & 0.84                    & 0.83                     & 0.83                    & 0.83                   & 0.87                    & 0.84                     & 0.83                    & 0.83                   & 0.97                    & 0.87                     & 0.86                    & 0.86                   \\

MDFEND \cite{zhu2022memory}      & 0.85                    &                     0.87     &      0.84                   & 0.85                   &    0.83                     &  0.82                        &           0.89              & 0.83                   & 0.95                    &   0.89                       &       0.87                  & 0.93                   \\
FuDFEND \cite{liang2022fudfend}  &        0.89                 &        0.88                  &   0.84                      & 0.90                   &    0.89                     &                0.87          &   0.84                      & 0.91                   &      0.96                   &   0.87                       &       0.80                  & 0.94                   \\
DITFEND \cite{nan2022improving}  & 0.85                    &                        0.82  &    0.80                     & 0.85                   &  0.86                       &  0.86                        &   0.83                      &    0.80                    & \textbf{0.97}                    &    0.86                      &   0.84                      & 0.94                   \\ 
SLFEND  \cite{wang2023soft}      &     0.91                    &       0.87                   &   0.84                      & \textbf{0.93}                   & \textbf{0.91}                        &                   \textbf{0.92}       &    0.90                     & 0.92                   &    0.92                     &  0.86                        &            0.85             & \textbf{0.95}                   \\
GPT-3.5 turbo (Prompting)                              & 0.73                    & 0.73                     & 0.73                    & 0.73                   & 0.61                    & 0.60                     & 0.60                    & 0.61                   & 0.62                    & 0.61                     & 0.61                    & 0.61                   \\
\hline

\textbf{\textit{CoALFake}}                                      &       \textbf{0.92}                  &    \textbf{0.91}                      &     \textbf{0.88}                    & \textbf{0.93}                       &         \textbf{0.91}                &  0.90                        &   \textbf{0.91}                      &  \textbf{0.93}                      &         0.96                & \textbf{0.91}                         &      \textbf{0.88}                   & \textbf{0.95}                       \\ 
\hline

Ablation Study                                      &                         &                          &                         &                        &                         &                          &                         &                        &                         &                          &                         &                        \\ 
(-) Cross-Attention                                      &       0.78                  &    0.77                      &  0.61                       & 0.85                       &      0.86                   &   0.86                       &        0.90                 &      0.90                  &    0.87                     &   0.81                       &      0.87                   &  0.90                      \\ 
(-) Domain-specific                                     &                   0.87      & 0.87                         &        0.83                 &    0.88                    &       0.89                  &     0.90                     &     0.89                 &  0.92                      &      0.92                   &  0.88                        &   0.87                      & 0.93                       \\ 
(-) Domain-shared                                     &      0.90                   &   0.89                       &      0.86                   &    0.91                    &   0.89                      &   0.89                       &       0.90                  &   0.92                     &     0.94                    &  0.89                        &   0.87                      &  0.93                      \\ 
 \bottomrule
\end{tabular}
\end{table}
 
For instance, on the Politifact dataset, \textit{CoALFake} achieves an F1 score of 0.92, significantly surpassing the next best baseline model, which scores 0.85. Likewise, \textit{CoALFake} achieves F1 scores of 0.91 and 0.95 on the Gossipcop and CoAID datasets, respectively, further underscoring its superior performance.
In contrast to models like FuDFEND \cite{liang2022fudfend} and SLFEND \cite{wang2023soft}, which use hard labels to represent the domain of a news record, our approach employs a vector to represent domain likelihood. This allows \textit{CoALFake} to accurately capture the probability of each record belonging to different domains. While EDDFN \cite{silva2021embracing} and MDFEND \cite{zhu2022memory} focus on integrating domain-specific and cross-domain knowledge, they may not fully leverage shared knowledge across domains or adapt as effectively to new domains. \textit{CoALFake}, however, excels in domain adaptation, with its dynamic attention mechanisms and ability to efficiently share knowledge across domains, contributing to its superior performance. These distinctions highlight the importance and effectiveness of our approach in comparison to the best baseline models.

A noteworthy observation throughout our experiments is that using GPT-3.5 Turbo (Prompting) to directly detect fake news underperforms compared to our model, which is trained on labels generated by GPT-3.5 Turbo.  These results partially correspond with prior findings in knowledge distillation \cite{wang2021zero} and pseudo-label-based learning \cite{min2022pseudo}, which, while sharing similar frameworks, differ slightly in their settings compared to \textit{CoALFake}.

\noindent\textbf{Ablation Study:}
The results from the ablation study emphasize the critical role of each feature in the model's overall performance. For example, when cross-attention is removed, the accuracy on the Gossipcop dataset drops from 0.91 to 0.86, demonstrating that this mechanism is vital for enhancing model accuracy. Likewise, removing domain-specific or domain-shared features results in a noticeable decrease in performance across all metrics, further underscoring the significance of these features in improving the model's robustness and effectiveness in fake news detection. Consequently, it is crucial to include a domain-specific layer to preserve domain-specific knowledge, a separate cross-domain layer for transferring common knowledge and a cross-attention mechanism to facilitate interaction between the two subspaces.
\subsection{Evaluation of Domain-Aware Sampling}
\begin{figure}[ht]
\centering
\includegraphics[width=\textwidth]{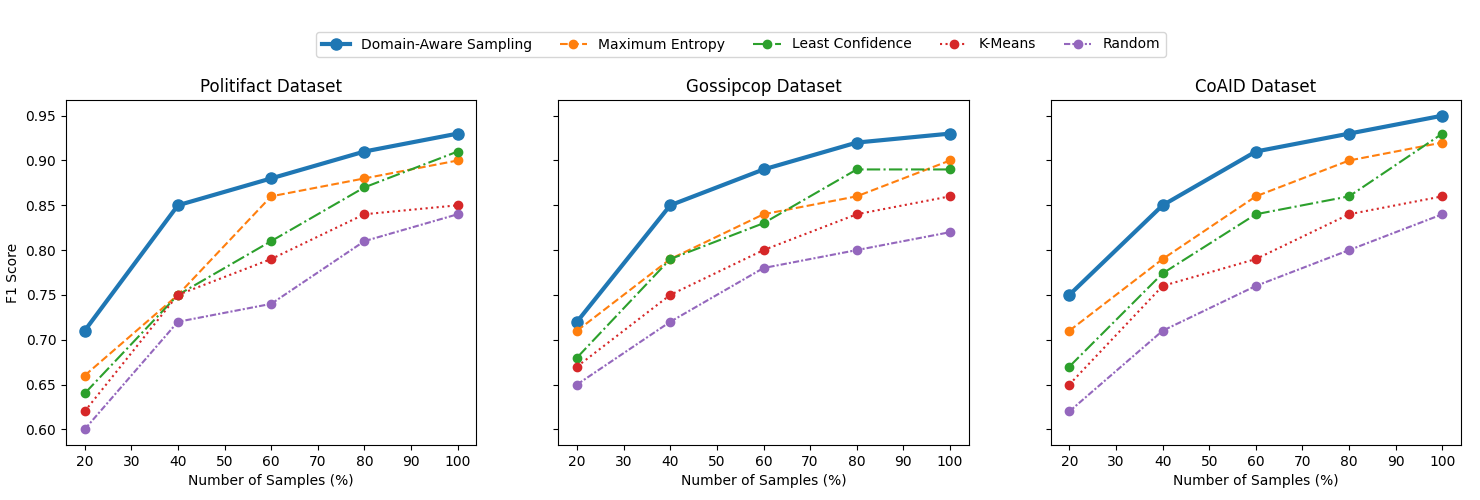}
\caption{\textit{CoALFake}’s performance with different sampling strategies. } 
\label{fig:sampling}
\end{figure}

Figure~\ref{fig:sampling} illustrates the performance of \textit{\textit{CoALFake}} with different active learning strategies (Section~\ref{sampling}) across all datasets. Among these methods, domain-aware sampling consistently outperforms the others, achieving the highest F1 scores at every sample percentage. In contrast, random sampling exhibits the lowest performance, with a slower and more gradual improvement in F1 scores. 

Uncertainty-based methods, i.e., maximal entropy and least confidence, significantly outperform the random baseline, demonstrating faster convergence and higher F1 scores by the end of the iterations. Although $k$-means clustering promotes diversity in feature space, its performance remains closely aligned with that of random sampling. Our proposed sampling strategy, which integrates both uncertainty and diversity, surpasses all other sampling strategies, achieving superior performance across datasets.
\subsection{Evaluation of Human-LLM Co-annotation}\label{sec:annotation}
In this section, we evaluate the effectiveness of our proposed co-annotation framework. We focus on two aspects: (1) how prompting strategies influence LLM inference quality and (2) how human re-annotation impacts model performance and labeling cost.
\subsubsection{Plain Instructions vs. $k$-NN Examples}
We access OpenAI APIs through the Azure service, utilizing GPT-3.5 Turbo as the LLM annotator for our experiments. The temperature is set to 0. Below the  prompt used for annotation:

\lstset{
  basicstyle=\ttfamily,
  breaklines=true
}

\begin{lstlisting}[language=Python, caption=Annotation Prompt.]
I need your assistance in evaluating the authenticity of a news article. 
I will provide you the news article. You have to answer only with Fake or Real. 
I will give you some examples of news. Your answer after [output] should be consistent with the following examples:

[example 1]: 
[input news]: [news text: {...}] 
[output]: [This is {...} news]

[example 2]: 
[input news]: [news text: {...}] 
[output]: [This is {...} news]

[target news]: 
[input news]: [news text: {...}]
[output]
\end{lstlisting}
Table~\ref{tab:knn} compares its inference performance using plain instructions versus when we add $k$-NN examples.
\begin{table}[ht]
\centering
\caption{Comparison Plain Instructions vs. Optimized Prompts.}\label{tab:knn}
\begin{tabular}{llll}
\toprule
                 & Politifact & Gossipcop & CoAID \\ \midrule
Base Instruction & 0.89       & 0.84      & 0.81  \\ 
+$k$-NN Examples   & \textbf{0.93}       & \textbf{0.93}      & \textbf{0.95}  \\ \bottomrule
\end{tabular}
\end{table}
The results show that optimized prompts significantly boost performance compared to plain instructions, emphasizing the value of tailored, context-rich prompts for handling challenging tasks.

\subsubsection{Cost vs. Performance Analysis}
In this section, we compare the labelling costs of GPT-3.5-turbo and crowdsourced labelling. For simplicity, we exclude costs related to GPT-3.5-turbo template selection, human labeler selection and other factors, focusing solely on the cost per label charged by the API or crowdsourcing platform.

The GPT-3.5-turbo API offered by OpenAI charges based on the number of tokens used for encoding and generation. According to OpenAI's pricing\footnote{https://platform.openai.com/docs/pricing}, the cost is \$3.00 for input and \$6.00 for output per 1 million tokens. Since the sequence length can vary significantly between different datasets, the cost of labelling a single instance with GPT-3.5-turbo also varies. Additionally, various few-shot labelling strategies with GPT-3.5-turbo incur different costs, with more shots leading to a higher labelling cost due to the longer prompt. For our experiments, we track the number of tokens used in each API call.

We estimate the crowdsourcing labelling price from Google Cloud Platform\footnote{https://cloud.google.com/ai-platform/data-labeling/pricing\#labeling\_costs}. For labeling classification tasks, it charges 1000 units (50 tokens per unit) for \$129 in Tier 1 and \$90 in Tier 2. We adopt the average cost from Tier 1 and Tier 2 as the human labelling cost. For generation tasks, there is no detailed instruction, as the rate can vary significantly based on task difficulty. Therefore, we follow the cost of classification tasks by charging \$0.11 per 50 tokens. It is important to note that actual human labelling is often more expensive. For example, the same instance is labelled by multiple labelers for majority voting and some datasets are labelled by experts rather than through crowdsourcing. 

\begin{figure}[h]
\centering
\includegraphics[width=\textwidth]{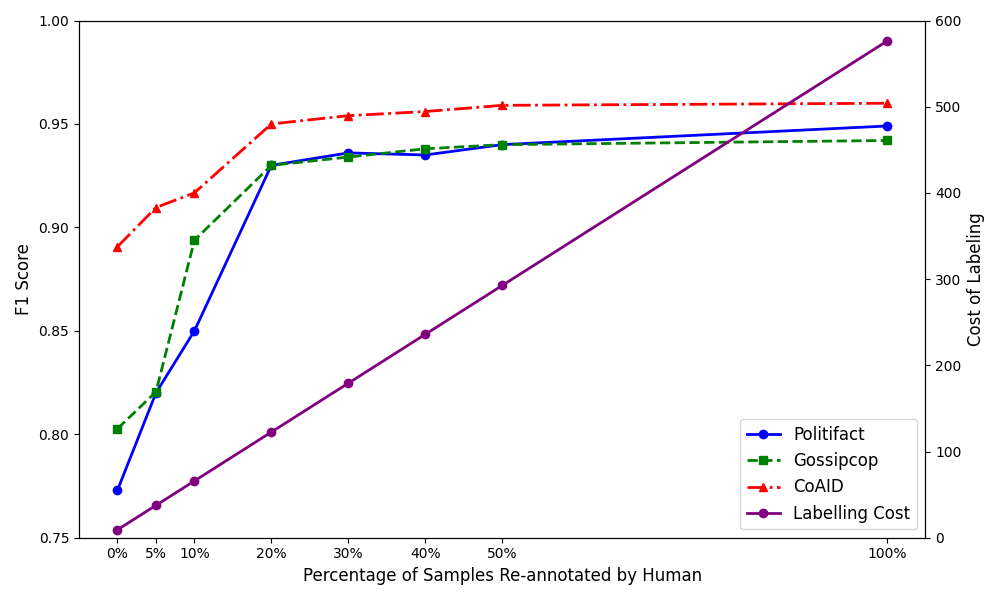}
\caption{F1 Score and Labelling Cost vs. Percentage of Human Re-annotation. } 
\label{fig:humanLLMCost}
\end{figure}

Figure~\ref{fig:humanLLMCost} illustrates the relationship between human re-annotation, model performance and labelling cost. 
As human involvement increases, F1 scores consistently improve across all three datasets. Without human re-annotation, Politifact, Gossipcop and CoAID achieve initial F1 scores of 0.773, 0.8027 and 0.8907, respectively. Even a small amount of human input (5–10\%) yields noticeable gains, while a more substantial increase to 20\% results in a significant performance boost, Politifact and Gossipcop reach approximately 0.93 and CoAID nears 0.95. Beyond this point, additional human annotations lead to diminishing returns, with full re-annotation achieving F1 scores of 0.949, 0.942 and 0.96.
However, these performance improvements come at a considerable cost. While fully automated labelling is highly economical at \$9.21, incorporating 20\% human re-annotation raises the cost substantially to around \$122.57. Beyond this threshold, the cost continues to escalate sharply, reaching nearly \$576.01 for full human annotation, while the corresponding performance improvements remain marginal.
These findings suggest that 20\% human re-annotation balances cost and performance effectively, surpassing baselines.

\subsubsection{Evaluation of Using LLM Labeling vs. LLM as Detector}
To assess the role of LLMs in \textit{CoALFake}, we compare two paradigms: (1) using GPT-3.5-Turbo directly as a detector and (2) using GPT-3.5-Turbo solely as an annotator to generate training labels for a downstream model.

As shown in Table~\ref{tab:results}, GPT-3.5-Turbo used in a prompting-based setting underperforms compared to a model trained on its generated labels within the \textit{CoALFake} pipeline. While this result may initially seem to benefit from the hybrid human–LLM annotation loop, we further isolate the effect by comparing it against the 0\% human re-annotation condition shown in Figure~\ref{fig:humanLLMCost}, where the model is trained using only raw GPT-3.5 annotations without human correction. Even in this purely LLM-labeled scenario, \textit{CoALFake} achieves higher performance than GPT-3.5 used directly as a detector. For example, on the PolitiFact dataset, GPT-3.5 achieves an F1 score of 0.73, whereas \textit{CoALFake} achieves 0.77. On GossipCop, the scores are 0.61 vs. 0.80, and on CoAID, 0.61 vs. 0.89.

These results highlight a key insight: LLMs are more effective as annotators than as standalone detectors. Training a domain-aware model on LLM-generated annotations allows for better generalization, particularly in cross-domain settings, supporting the design choice of incorporating LLMs into a human-in-the-loop learning framework rather than relying on them for direct classification.
\section{Discussion}
In this section, we present a comparative discussion of \textit{CoALFake} in the context of existing human–AI annotation frameworks, followed by a reflection on its limitations and the broader societal and ethical implications.
\subsection{Beyond Pre-Annotation: Comparing \textit{CoALFake} with Human–AI Annotation Frameworks}
While human-AI collaborative annotation has been widely explored, most existing frameworks adopt a simplistic model in which machine-generated annotations are either directly validated by humans or loosely combined in ensemble schemes. A commonly used approach is automatic pre-annotation followed by human correction~\cite{skeppstedt2013annotating}, which has shown promise in reducing annotation time, but only when pre-annotations are highly accurate~\cite{mikulova2023quality}. In contrast, low-quality pre-annotations have been found to yield degraded label quality~\cite{south2014evaluating}.

\textit{CoALFake} advances beyond this baseline. Instead of relying on static or opaque ML pre-annotators, it leverages the adaptive capabilities of LLMs. Moreover, it integrates a label verification mechanism using CL to identify likely label errors, thereby avoiding the inefficiencies of blanket human review. This design ensures that human expertise is focused where it is most needed, resulting in a much more efficient and reliable annotation pipeline.

Unlike prior frameworks that treat human and AI contributions as interchangeable or apply human validation uniformly, \textit{CoALFake} establishes a principled and asymmetrical collaboration model: LLMs handle the scalable annotation workload, while humans provide strategic oversight. This structured division of labor not only enhances annotation quality but also significantly reduces manual effort.

\subsection{Limitations}
Although \textit{CoALFake} demonstrates strong empirical performance, several limitations remain. First, the framework relies on the original dataset labels as a proxy for human re-annotations. This assumption may limit generalizability across different platforms, contexts, or expert annotation standards. Second, the effectiveness of our label verification step can vary depending on task complexity and model calibration. Third, the performance of the LLM annotator (GPT-3.5-Turbo) plays a central role in the quality of \textit{CoALFake}'s labels. Due to resource constraints, we did not evaluate other LLMs. Future work should assess the generalizability of \textit{CoALFake} across different LLMs. Fourth, while \textit{CoALFake} is evaluated across multiple domains, its ability to generalize to entirely unseen domains remains untested. A leave-one-domain-out evaluation setting or testing on external datasets would better validate cross-domain robustness. Finally, the human re-annotation process may be improved by providing LLM-generated explanations. 

\subsection{Societal and Ethical Implications}
The use of LLMs for fake news detection introduces a range of societal and ethical considerations that go beyond accuracy. While \textit{CoALFake} integrates human oversight to mitigate the risks of erroneous AI-generated annotations, it is important to acknowledge broader concerns around human–machine collaboration in high-stakes domains like news and media.

First, \textit{CoALFake} is designed around selective human verification, where LLMs serve as scalable annotators, and humans intervene only in cases identified as likely mislabeled. This asymmetrical structure still values human judgment in critical decision points. However, this paradigm also raises concerns about the diminishing role of human expertise and increased reliance on algorithmic outputs. Second, the presence of LLMs in the fake news detection pipeline may inadvertently shape public discourse, especially when used to annotate large-scale content. If unchecked, LLMs may encode and reinforce latent biases present in training data, leading to biased labeling of politically or culturally sensitive content. While \textit{CoALFake} mitigates this by incorporating human re-annotation and uncertainty estimation, broader deployments would require transparent auditing mechanisms and bias detection tools. Third, there are legal and regulatory issues around using commercial LLMs for annotation, especially when handling sensitive or personally identifiable content. Ensuring data privacy and model accountability is essential for responsible deployment.
\section{Conclusion and Future Work}
This paper introduces \textit{CoALFake}, a novel framework for cross-domain fake news detection that integrates Human–LLM co-annotation with domain-aware active learning. The proposed approach reduces human annotation costs while improving generalization across diverse domains. By leveraging both domain-specific and cross-domain representations, \textit{CoALFake} demonstrates consistent performance improvements over existing baselines.

Future work may extend this system in several directions. First, multimodal detection can be explored by incorporating images, metadata, or user interaction features, which may enhance the robustness of fake news detection. Second, given the rapid advancement of LLMs, future iterations may benefit from employing next-generation LLMs. An ensemble-based annotation strategy, using multiple LLMs with majority voting, could further improve label quality. Additionally, the quality of few-shot prompting could be enhanced by investigating task-specific or adaptive prompt design. Finally, the human re-annotation process could be facilitated by providing LLM-generated explanations; carefully designed explanations may assist annotators in resolving ambiguous cases and improving annotation consistency.

\bibliography{sn-bibliography}


\begin{thebibliography}{45}
\ifx \bisbn   \undefined \def \bisbn  #1{ISBN #1}\fi
\ifx \binits  \undefined \def \binits#1{#1}\fi
\ifx \bauthor  \undefined \def \bauthor#1{#1}\fi
\ifx \batitle  \undefined \def \batitle#1{#1}\fi
\ifx \bjtitle  \undefined \def \bjtitle#1{#1}\fi
\ifx \bvolume  \undefined \def \bvolume#1{\textbf{#1}}\fi
\ifx \byear  \undefined \def \byear#1{#1}\fi
\ifx \bissue  \undefined \def \bissue#1{#1}\fi
\ifx \bfpage  \undefined \def \bfpage#1{#1}\fi
\ifx \blpage  \undefined \def \blpage #1{#1}\fi
\ifx \burl  \undefined \def \burl#1{\textsf{#1}}\fi
\ifx \doiurl  \undefined \def \doiurl#1{\url{https://doi.org/#1}}\fi
\ifx \betal  \undefined \def \betal{\textit{et al.}}\fi
\ifx \binstitute  \undefined \def \binstitute#1{#1}\fi
\ifx \binstitutionaled  \undefined \def \binstitutionaled#1{#1}\fi
\ifx \bctitle  \undefined \def \bctitle#1{#1}\fi
\ifx \beditor  \undefined \def \beditor#1{#1}\fi
\ifx \bpublisher  \undefined \def \bpublisher#1{#1}\fi
\ifx \bbtitle  \undefined \def \bbtitle#1{#1}\fi
\ifx \bedition  \undefined \def \bedition#1{#1}\fi
\ifx \bseriesno  \undefined \def \bseriesno#1{#1}\fi
\ifx \blocation  \undefined \def \blocation#1{#1}\fi
\ifx \bsertitle  \undefined \def \bsertitle#1{#1}\fi
\ifx \bsnm \undefined \def \bsnm#1{#1}\fi
\ifx \bsuffix \undefined \def \bsuffix#1{#1}\fi
\ifx \bparticle \undefined \def \bparticle#1{#1}\fi
\ifx \barticle \undefined \def \barticle#1{#1}\fi
\bibcommenthead
\ifx \bconfdate \undefined \def \bconfdate #1{#1}\fi
\ifx \botherref \undefined \def \botherref #1{#1}\fi
\ifx \url \undefined \def \url#1{\textsf{#1}}\fi
\ifx \bchapter \undefined \def \bchapter#1{#1}\fi
\ifx \bbook \undefined \def \bbook#1{#1}\fi
\ifx \bcomment \undefined \def \bcomment#1{#1}\fi
\ifx \oauthor \undefined \def \oauthor#1{#1}\fi
\ifx \citeauthoryear \undefined \def \citeauthoryear#1{#1}\fi
\ifx \endbibitem  \undefined \def \endbibitem {}\fi
\ifx \bconflocation  \undefined \def \bconflocation#1{#1}\fi
\ifx \arxivurl  \undefined \def \arxivurl#1{\textsf{#1}}\fi
\csname PreBibitemsHook\endcsname

%%% 1
\bibitem[\protect\citeauthoryear{Sallami et~al.}{2023}]{sallami2023hype}
\begin{botherref}
\oauthor{\bsnm{Sallami}, \binits{D.}},
\oauthor{\bsnm{Gueddiche}, \binits{A.}},
\oauthor{\bsnm{A{\"\i}meur}, \binits{E.}}:
From hype to reality: Revealing the accuracy and robustness of transformer-based models for fake news detection
(2023)
\end{botherref}
\endbibitem

%%% 2
\bibitem[\protect\citeauthoryear{Sallami and A{\"\i}meur}{2025}]{sallami2025exploring}
\begin{barticle}
\bauthor{\bsnm{Sallami}, \binits{D.}},
\bauthor{\bsnm{A{\"\i}meur}, \binits{E.}}:
\batitle{Exploring beyond detection: a review on fake news prevention and mitigation techniques}.
\bjtitle{Journal of Computational Social Science}
\bvolume{8}(\bissue{1}),
\bfpage{1}--\blpage{38}
(\byear{2025})
\end{barticle}
\endbibitem

%%% 3
\bibitem[\protect\citeauthoryear{Li et~al.}{2023}]{li2023improving}
\begin{bchapter}
\bauthor{\bsnm{Li}, \binits{J.}},
\bauthor{\bsnm{Wang}, \binits{L.}},
\bauthor{\bsnm{He}, \binits{J.}},
\bauthor{\bsnm{Zhang}, \binits{Y.}},
\bauthor{\bsnm{Liu}, \binits{A.}}:
\bctitle{Improving rumor detection by class-based adversarial domain adaptation}.
In: \bbtitle{Proceedings of the 31st ACM International Conference on Multimedia},
pp. \bfpage{6634}--\blpage{6642}
(\byear{2023})
\end{bchapter}
\endbibitem

%%% 4
\bibitem[\protect\citeauthoryear{A{\"\i}meur et~al.}{2025}]{aimeur2025too}
\begin{bchapter}
\bauthor{\bsnm{A{\"\i}meur}, \binits{E.}},
\bauthor{\bsnm{Brassard}, \binits{G.}},
\bauthor{\bsnm{Sallami}, \binits{D.}}:
\bctitle{Too focused on accuracy to notice the fallout: Towards socially responsible fake news detection}.
In: \bbtitle{Proceedings of the AAAI/ACM Conference on AI, Ethics, and Society},
vol. \bseriesno{8},
pp. \bfpage{55}--\blpage{65}
(\byear{2025})
\end{bchapter}
\endbibitem

%%% 5
\bibitem[\protect\citeauthoryear{Silva et~al.}{2021}]{silva2021embracing}
\begin{bchapter}
\bauthor{\bsnm{Silva}, \binits{A.}},
\bauthor{\bsnm{Luo}, \binits{L.}},
\bauthor{\bsnm{Karunasekera}, \binits{S.}},
\bauthor{\bsnm{Leckie}, \binits{C.}}:
\bctitle{Embracing domain differences in fake news: Cross-domain fake news detection using multi-modal data}.
In: \bbtitle{Proceedings of the AAAI Conference on Artificial Intelligence},
vol. \bseriesno{35},
pp. \bfpage{557}--\blpage{565}
(\byear{2021})
\end{bchapter}
\endbibitem

%%% 6
\bibitem[\protect\citeauthoryear{Wang et~al.}{2023}]{wang2023soft}
\begin{botherref}
\oauthor{\bsnm{Wang}, \binits{D.}},
\oauthor{\bsnm{Zhang}, \binits{W.}},
\oauthor{\bsnm{Wu}, \binits{W.}},
\oauthor{\bsnm{Guo}, \binits{X.}}:
Soft-label for multi-domain fake news detection.
IEEE Access
(2023)
\end{botherref}
\endbibitem

%%% 7
\bibitem[\protect\citeauthoryear{Lin et~al.}{2022}]{lin2022detect}
\begin{botherref}
\oauthor{\bsnm{Lin}, \binits{H.}},
\oauthor{\bsnm{Ma}, \binits{J.}},
\oauthor{\bsnm{Chen}, \binits{L.}},
\oauthor{\bsnm{Yang}, \binits{Z.}},
\oauthor{\bsnm{Cheng}, \binits{M.}},
\oauthor{\bsnm{Chen}, \binits{G.}}:
Detect rumors in microblog posts for low-resource domains via adversarial contrastive learning.
arXiv preprint arXiv:2204.08143
(2022)
\end{botherref}
\endbibitem

%%% 8
\bibitem[\protect\citeauthoryear{Lu et~al.}{2021}]{lu2021fantastically}
\begin{botherref}
\oauthor{\bsnm{Lu}, \binits{Y.}},
\oauthor{\bsnm{Bartolo}, \binits{M.}},
\oauthor{\bsnm{Moore}, \binits{A.}},
\oauthor{\bsnm{Riedel}, \binits{S.}},
\oauthor{\bsnm{Stenetorp}, \binits{P.}}:
Fantastically ordered prompts and where to find them: Overcoming few-shot prompt order sensitivity.
arXiv preprint arXiv:2104.08786
(2021)
\end{botherref}
\endbibitem

%%% 9
\bibitem[\protect\citeauthoryear{Ma et~al.}{2024}]{ma2024fake}
\begin{bchapter}
\bauthor{\bsnm{Ma}, \binits{X.}},
\bauthor{\bsnm{Zhang}, \binits{Y.}},
\bauthor{\bsnm{Ding}, \binits{K.}},
\bauthor{\bsnm{Yang}, \binits{J.}},
\bauthor{\bsnm{Wu}, \binits{J.}},
\bauthor{\bsnm{Fan}, \binits{H.}}:
\bctitle{On fake news detection with llm enhanced semantics mining}.
In: \bbtitle{Proceedings of the 2024 Conference on Empirical Methods in Natural Language Processing},
pp. \bfpage{508}--\blpage{521}
(\byear{2024})
\end{bchapter}
\endbibitem

%%% 10
\bibitem[\protect\citeauthoryear{Wang et~al.}{2025}]{wang2025fake}
\begin{botherref}
\oauthor{\bsnm{Wang}, \binits{X.}},
\oauthor{\bsnm{Meng}, \binits{J.}},
\oauthor{\bsnm{Zhao}, \binits{D.}},
\oauthor{\bsnm{Meng}, \binits{X.}},
\oauthor{\bsnm{Sun}, \binits{H.}}:
Fake news detection based on multi-modal domain adaptation.
Neural Computing and Applications,
1--13
(2025)
\end{botherref}
\endbibitem

%%% 11
\bibitem[\protect\citeauthoryear{Han et~al.}{2020}]{han2020graph}
\begin{botherref}
\oauthor{\bsnm{Han}, \binits{Y.}},
\oauthor{\bsnm{Karunasekera}, \binits{S.}},
\oauthor{\bsnm{Leckie}, \binits{C.}}:
Graph neural networks with continual learning for fake news detection from social media.
arXiv preprint arXiv:2007.03316
(2020)
\end{botherref}
\endbibitem

%%% 12
\bibitem[\protect\citeauthoryear{Nan et~al.}{2021}]{nan2021mdfend}
\begin{bchapter}
\bauthor{\bsnm{Nan}, \binits{Q.}},
\bauthor{\bsnm{Cao}, \binits{J.}},
\bauthor{\bsnm{Zhu}, \binits{Y.}},
\bauthor{\bsnm{Wang}, \binits{Y.}},
\bauthor{\bsnm{Li}, \binits{J.}}:
\bctitle{Mdfend: Multi-domain fake news detection}.
In: \bbtitle{Proceedings of the 30th ACM International Conference on Information \& Knowledge Management},
pp. \bfpage{3343}--\blpage{3347}
(\byear{2021})
\end{bchapter}
\endbibitem

%%% 13
\bibitem[\protect\citeauthoryear{Liang et~al.}{2022}]{liang2022fudfend}
\begin{bchapter}
\bauthor{\bsnm{Liang}, \binits{C.}},
\bauthor{\bsnm{Zhang}, \binits{Y.}},
\bauthor{\bsnm{Li}, \binits{X.}},
\bauthor{\bsnm{Zhang}, \binits{J.}},
\bauthor{\bsnm{Yu}, \binits{Y.}}:
\bctitle{Fudfend: fuzzy-domain for multi-domain fake news detection}.
In: \bbtitle{CCF International Conference on Natural Language Processing and Chinese Computing},
pp. \bfpage{45}--\blpage{57}
(\byear{2022}).
\bcomment{Springer}
\end{bchapter}
\endbibitem

%%% 14
\bibitem[\protect\citeauthoryear{Rastogi et~al.}{2021}]{rastogi2021adaptive}
\begin{bchapter}
\bauthor{\bsnm{Rastogi}, \binits{S.}},
\bauthor{\bsnm{Gill}, \binits{S.S.}},
\bauthor{\bsnm{Bansal}, \binits{D.}}:
\bctitle{An adaptive approach for fake news detection in social media: single vs cross domain}.
In: \bbtitle{2021 International Conference on Computational Science and Computational Intelligence (CSCI)},
pp. \bfpage{1401}--\blpage{1405}
(\byear{2021}).
\bcomment{IEEE}
\end{bchapter}
\endbibitem

%%% 15
\bibitem[\protect\citeauthoryear{Amri et~al.}{2021}]{amri2021exmulf}
\begin{bchapter}
\bauthor{\bsnm{Amri}, \binits{S.}},
\bauthor{\bsnm{Sallami}, \binits{D.}},
\bauthor{\bsnm{A{\"\i}meur}, \binits{E.}}:
\bctitle{Exmulf: An explainable multimodal content-based fake news detection system}.
In: \bbtitle{International Symposium on Foundations and Practice of Security},
pp. \bfpage{177}--\blpage{187}
(\byear{2021}).
\bcomment{Springer}
\end{bchapter}
\endbibitem

%%% 16
\bibitem[\protect\citeauthoryear{Wang et~al.}{2020}]{wang2020weak}
\begin{bchapter}
\bauthor{\bsnm{Wang}, \binits{Y.}},
\bauthor{\bsnm{Yang}, \binits{W.}},
\bauthor{\bsnm{Ma}, \binits{F.}},
\bauthor{\bsnm{Xu}, \binits{J.}},
\bauthor{\bsnm{Zhong}, \binits{B.}},
\bauthor{\bsnm{Deng}, \binits{Q.}},
\bauthor{\bsnm{Gao}, \binits{J.}}:
\bctitle{Weak supervision for fake news detection via reinforcement learning}.
In: \bbtitle{Proceedings of the AAAI Conference on Artificial Intelligence},
vol. \bseriesno{34},
pp. \bfpage{516}--\blpage{523}
(\byear{2020})
\end{bchapter}
\endbibitem

%%% 17
\bibitem[\protect\citeauthoryear{Ren et~al.}{2020}]{ren2020adversarial}
\begin{bchapter}
\bauthor{\bsnm{Ren}, \binits{Y.}},
\bauthor{\bsnm{Wang}, \binits{B.}},
\bauthor{\bsnm{Zhang}, \binits{J.}},
\bauthor{\bsnm{Chang}, \binits{Y.}}:
\bctitle{Adversarial active learning based heterogeneous graph neural network for fake news detection}.
In: \bbtitle{2020 IEEE International Conference on Data Mining (ICDM)},
pp. \bfpage{452}--\blpage{461}
(\byear{2020}).
\bcomment{IEEE}
\end{bchapter}
\endbibitem

%%% 18
\bibitem[\protect\citeauthoryear{Barnab{\`o} et~al.}{2023}]{barnabo2023deep}
\begin{barticle}
\bauthor{\bsnm{Barnab{\`o}}, \binits{G.}},
\bauthor{\bsnm{Siciliano}, \binits{F.}},
\bauthor{\bsnm{Castillo}, \binits{C.}},
\bauthor{\bsnm{Leonardi}, \binits{S.}},
\bauthor{\bsnm{Nakov}, \binits{P.}},
\bauthor{\bsnm{Da~San~Martino}, \binits{G.}},
\bauthor{\bsnm{Silvestri}, \binits{F.}}:
\batitle{Deep active learning for misinformation detection using geometric deep learning}.
\bjtitle{Online Social Networks and Media}
\bvolume{33},
\bfpage{100244}
(\byear{2023})
\end{barticle}
\endbibitem

%%% 19
\bibitem[\protect\citeauthoryear{Folino et~al.}{2024}]{folino2024towards}
\begin{barticle}
\bauthor{\bsnm{Folino}, \binits{F.}},
\bauthor{\bsnm{Folino}, \binits{G.}},
\bauthor{\bsnm{Guarascio}, \binits{M.}},
\bauthor{\bsnm{Pontieri}, \binits{L.}},
\bauthor{\bsnm{Zicari}, \binits{P.}}:
\batitle{Towards data-and compute-efficient fake-news detection: An approach combining active learning and pre-trained language models}.
\bjtitle{SN Computer Science}
\bvolume{5}(\bissue{5}),
\bfpage{470}
(\byear{2024})
\end{barticle}
\endbibitem

%%% 20
\bibitem[\protect\citeauthoryear{Xiao et~al.}{}]{xiao2023freeal}
\begin{botherref}
\oauthor{\bsnm{Xiao}, \binits{R.}},
\oauthor{\bsnm{Dong}, \binits{Y.}},
\oauthor{\bsnm{Zhao}, \binits{J.}},
\oauthor{\bsnm{Wu}, \binits{R.}},
\oauthor{\bsnm{Lin}, \binits{M.}},
\oauthor{\bsnm{Chen}, \binits{G.}},
\oauthor{\bsnm{Wang}, \binits{H.}}:
Freeal: Towards human-free active learning in the era of large language models.
In: The 2023 Conference on Empirical Methods in Natural Language Processing
\end{botherref}
\endbibitem

%%% 21
\bibitem[\protect\citeauthoryear{Zhang et~al.}{2023}]{zhang2023llmaaa}
\begin{bchapter}
\bauthor{\bsnm{Zhang}, \binits{R.}},
\bauthor{\bsnm{Li}, \binits{Y.}},
\bauthor{\bsnm{Ma}, \binits{Y.}},
\bauthor{\bsnm{Zhou}, \binits{M.}},
\bauthor{\bsnm{Zou}, \binits{L.}}:
\bctitle{Llmaaa: Making large language models as active annotators}.
In: \bbtitle{Findings of the Association for Computational Linguistics: EMNLP 2023},
pp. \bfpage{13088}--\blpage{13103}
(\byear{2023})
\end{bchapter}
\endbibitem

%%% 22
\bibitem[\protect\citeauthoryear{Shahapure and Nicholas}{2020}]{shahapure2020cluster}
\begin{bchapter}
\bauthor{\bsnm{Shahapure}, \binits{K.R.}},
\bauthor{\bsnm{Nicholas}, \binits{C.}}:
\bctitle{Cluster quality analysis using silhouette score}.
In: \bbtitle{2020 IEEE 7th International Conference on Data Science and Advanced Analytics (DSAA)},
pp. \bfpage{747}--\blpage{748}
(\byear{2020}).
\bcomment{IEEE}
\end{bchapter}
\endbibitem

%%% 23
\bibitem[\protect\citeauthoryear{Brown et~al.}{2020}]{brown2020language}
\begin{barticle}
\bauthor{\bsnm{Brown}, \binits{T.}},
\bauthor{\bsnm{Mann}, \binits{B.}},
\bauthor{\bsnm{Ryder}, \binits{N.}},
\bauthor{\bsnm{Subbiah}, \binits{M.}},
\bauthor{\bsnm{Kaplan}, \binits{J.D.}},
\bauthor{\bsnm{Dhariwal}, \binits{P.}},
\bauthor{\bsnm{Neelakantan}, \binits{A.}},
\bauthor{\bsnm{Shyam}, \binits{P.}},
\bauthor{\bsnm{Sastry}, \binits{G.}},
\bauthor{\bsnm{Askell}, \binits{A.}}, \betal:
\batitle{Language models are few-shot learners}.
\bjtitle{Advances in neural information processing systems}
\bvolume{33},
\bfpage{1877}--\blpage{1901}
(\byear{2020})
\end{barticle}
\endbibitem

%%% 24
\bibitem[\protect\citeauthoryear{Liu et~al.}{2021}]{liu2021makes}
\begin{botherref}
\oauthor{\bsnm{Liu}, \binits{J.}},
\oauthor{\bsnm{Shen}, \binits{D.}},
\oauthor{\bsnm{Zhang}, \binits{Y.}},
\oauthor{\bsnm{Dolan}, \binits{B.}},
\oauthor{\bsnm{Carin}, \binits{L.}},
\oauthor{\bsnm{Chen}, \binits{W.}}:
What makes good in-context examples for gpt-$3 $?
arXiv preprint arXiv:2101.06804
(2021)
\end{botherref}
\endbibitem

%%% 25
\bibitem[\protect\citeauthoryear{Reimers}{2019}]{reimers2019sentence}
\begin{botherref}
\oauthor{\bsnm{Reimers}, \binits{N.}}:
Sentence-bert: Sentence embeddings using siamese bert-networks.
arXiv preprint arXiv:1908.10084
(2019)
\end{botherref}
\endbibitem

%%% 26
\bibitem[\protect\citeauthoryear{Lin et~al.}{2022}]{lin2022teaching}
\begin{botherref}
\oauthor{\bsnm{Lin}, \binits{S.}},
\oauthor{\bsnm{Hilton}, \binits{J.}},
\oauthor{\bsnm{Evans}, \binits{O.}}:
Teaching models to express their uncertainty in words.
Transactions on Machine Learning Research
(2022)
\end{botherref}
\endbibitem

%%% 27
\bibitem[\protect\citeauthoryear{Wang et~al.}{2022}]{wangself}
\begin{bchapter}
\bauthor{\bsnm{Wang}, \binits{X.}},
\bauthor{\bsnm{Wei}, \binits{J.}},
\bauthor{\bsnm{Schuurmans}, \binits{D.}},
\bauthor{\bsnm{Le}, \binits{Q.V.}},
\bauthor{\bsnm{Chi}, \binits{E.H.}},
\bauthor{\bsnm{Narang}, \binits{S.}},
\bauthor{\bsnm{Chowdhery}, \binits{A.}},
\bauthor{\bsnm{Zhou}, \binits{D.}}:
\bctitle{Self-consistency improves chain of thought reasoning in language models}.
In: \bbtitle{The Eleventh International Conference on Learning Representations}
(\byear{2022})
\end{bchapter}
\endbibitem

%%% 28
\bibitem[\protect\citeauthoryear{Xiong et~al.}{2023}]{xiongcan}
\begin{bchapter}
\bauthor{\bsnm{Xiong}, \binits{M.}},
\bauthor{\bsnm{Hu}, \binits{Z.}},
\bauthor{\bsnm{Lu}, \binits{X.}},
\bauthor{\bsnm{LI}, \binits{Y.}},
\bauthor{\bsnm{Fu}, \binits{J.}},
\bauthor{\bsnm{He}, \binits{J.}},
\bauthor{\bsnm{Hooi}, \binits{B.}}:
\bctitle{Can llms express their uncertainty? an empirical evaluation of confidence elicitation in llms}.
In: \bbtitle{The Twelfth International Conference on Learning Representations}
(\byear{2023})
\end{bchapter}
\endbibitem

%%% 29
\bibitem[\protect\citeauthoryear{Northcutt et~al.}{2021}]{northcutt2021confident}
\begin{barticle}
\bauthor{\bsnm{Northcutt}, \binits{C.}},
\bauthor{\bsnm{Jiang}, \binits{L.}},
\bauthor{\bsnm{Chuang}, \binits{I.}}:
\batitle{Confident learning: Estimating uncertainty in dataset labels}.
\bjtitle{Journal of Artificial Intelligence Research}
\bvolume{70},
\bfpage{1373}--\blpage{1411}
(\byear{2021})
\end{barticle}
\endbibitem

%%% 30
\bibitem[\protect\citeauthoryear{Shu et~al.}{2020}]{shu2020fakenewsnet}
\begin{barticle}
\bauthor{\bsnm{Shu}, \binits{K.}},
\bauthor{\bsnm{Mahudeswaran}, \binits{D.}},
\bauthor{\bsnm{Wang}, \binits{S.}},
\bauthor{\bsnm{Lee}, \binits{D.}},
\bauthor{\bsnm{Liu}, \binits{H.}}:
\batitle{Fakenewsnet: A data repository with news content, social context, and spatiotemporal information for studying fake news on social media}.
\bjtitle{Big data}
\bvolume{8}(\bissue{3}),
\bfpage{171}--\blpage{188}
(\byear{2020})
\end{barticle}
\endbibitem

%%% 31
\bibitem[\protect\citeauthoryear{Cui and Lee}{2020}]{cui2020coaid}
\begin{botherref}
\oauthor{\bsnm{Cui}, \binits{L.}},
\oauthor{\bsnm{Lee}, \binits{D.}}:
Coaid: Covid-19 healthcare misinformation dataset.
arXiv preprint arXiv:2006.00885
(2020)
\end{botherref}
\endbibitem

%%% 32
\bibitem[\protect\citeauthoryear{Shu et~al.}{2019}]{shu2019defend}
\begin{bchapter}
\bauthor{\bsnm{Shu}, \binits{K.}},
\bauthor{\bsnm{Cui}, \binits{L.}},
\bauthor{\bsnm{Wang}, \binits{S.}},
\bauthor{\bsnm{Lee}, \binits{D.}},
\bauthor{\bsnm{Liu}, \binits{H.}}:
\bctitle{defend: Explainable fake news detection}.
In: \bbtitle{Proceedings of the 25th ACM SIGKDD International Conference on Knowledge Discovery \& Data Mining},
pp. \bfpage{395}--\blpage{405}
(\byear{2019})
\end{bchapter}
\endbibitem

%%% 33
\bibitem[\protect\citeauthoryear{Silva et~al.}{2020}]{silva2020embedding}
\begin{bchapter}
\bauthor{\bsnm{Silva}, \binits{A.}},
\bauthor{\bsnm{Han}, \binits{Y.}},
\bauthor{\bsnm{Luo}, \binits{L.}},
\bauthor{\bsnm{Karunasekera}, \binits{S.}},
\bauthor{\bsnm{Leckie}, \binits{C.}}:
\bctitle{Embedding partial propagation network for fake news early detection.}
In: \bbtitle{CIKM (Workshops)},
vol. \bseriesno{2699}
(\byear{2020})
\end{bchapter}
\endbibitem

%%% 34
\bibitem[\protect\citeauthoryear{Zhou et~al.}{2020}]{zhou2020similarity}
\begin{bchapter}
\bauthor{\bsnm{Zhou}, \binits{X.}},
\bauthor{\bsnm{Wu}, \binits{J.}},
\bauthor{\bsnm{Zafarani}, \binits{R.}}:
\bctitle{: Similarity-aware multi-modal fake news detection}.
In: \bbtitle{Pacific-Asia Conference on Knowledge Discovery and Data Mining},
pp. \bfpage{354}--\blpage{367}
(\byear{2020}).
\bcomment{Springer}
\end{bchapter}
\endbibitem

%%% 35
\bibitem[\protect\citeauthoryear{Zhu et~al.}{2022}]{zhu2022memory}
\begin{barticle}
\bauthor{\bsnm{Zhu}, \binits{Y.}},
\bauthor{\bsnm{Sheng}, \binits{Q.}},
\bauthor{\bsnm{Cao}, \binits{J.}},
\bauthor{\bsnm{Nan}, \binits{Q.}},
\bauthor{\bsnm{Shu}, \binits{K.}},
\bauthor{\bsnm{Wu}, \binits{M.}},
\bauthor{\bsnm{Wang}, \binits{J.}},
\bauthor{\bsnm{Zhuang}, \binits{F.}}:
\batitle{Memory-guided multi-view multi-domain fake news detection}.
\bjtitle{IEEE Transactions on Knowledge and Data Engineering}
\bvolume{35}(\bissue{7}),
\bfpage{7178}--\blpage{7191}
(\byear{2022})
\end{barticle}
\endbibitem

%%% 36
\bibitem[\protect\citeauthoryear{Nan et~al.}{2022}]{nan2022improving}
\begin{bchapter}
\bauthor{\bsnm{Nan}, \binits{Q.}},
\bauthor{\bsnm{Wang}, \binits{D.}},
\bauthor{\bsnm{Zhu}, \binits{Y.}},
\bauthor{\bsnm{Sheng}, \binits{Q.}},
\bauthor{\bsnm{Shi}, \binits{Y.}},
\bauthor{\bsnm{Cao}, \binits{J.}},
\bauthor{\bsnm{Li}, \binits{J.}}:
\bctitle{Improving fake news detection of influential domain via domain-and instance-level transfer}.
In: \bbtitle{Proceedings of the 29th International Conference on Computational Linguistics},
pp. \bfpage{2834}--\blpage{2848}
(\byear{2022})
\end{bchapter}
\endbibitem

%%% 37
\bibitem[\protect\citeauthoryear{Settles}{2009}]{settles2009active}
\begin{botherref}
\oauthor{\bsnm{Settles}, \binits{B.}}:
Active learning literature survey
(2009)
\end{botherref}
\endbibitem

%%% 38
\bibitem[\protect\citeauthoryear{Culotta and McCallum}{2005}]{culotta2005reducing}
\begin{bchapter}
\bauthor{\bsnm{Culotta}, \binits{A.}},
\bauthor{\bsnm{McCallum}, \binits{A.}}:
\bctitle{Reducing labeling effort for structured prediction tasks}.
In: \bbtitle{AAAI},
vol. \bseriesno{5},
pp. \bfpage{746}--\blpage{751}
(\byear{2005})
\end{bchapter}
\endbibitem

%%% 39
\bibitem[\protect\citeauthoryear{Yuan et~al.}{2020}]{yuan2020cold}
\begin{bchapter}
\bauthor{\bsnm{Yuan}, \binits{M.}},
\bauthor{\bsnm{Lin}, \binits{H.-T.}},
\bauthor{\bsnm{Boyd-Graber}, \binits{J.}}:
\bctitle{Cold-start active learning through self-supervised language modeling}.
In: \bbtitle{Proceedings of the 2020 Conference on Empirical Methods in Natural Language Processing (EMNLP)},
pp. \bfpage{7935}--\blpage{7948}
(\byear{2020})
\end{bchapter}
\endbibitem

%%% 40
\bibitem[\protect\citeauthoryear{Joulin et~al.}{2017}]{joulin2017bag}
\begin{bchapter}
\bauthor{\bsnm{Joulin}, \binits{A.}},
\bauthor{\bsnm{Grave}, \binits{E.}},
\bauthor{\bsnm{Bojanowski}, \binits{P.}},
\bauthor{\bsnm{Mikolov}, \binits{T.}}:
\bctitle{Bag of tricks for efficient text classification}.
In: \bbtitle{Proceedings of the 15th Conference of the European Chapter of the Association for Computational Linguistics: Volume 2, Short Papers}
(\byear{2017}).
\bcomment{Association for Computational Linguistics}
\end{bchapter}
\endbibitem

%%% 41
\bibitem[\protect\citeauthoryear{Wang}{2021}]{wang2021zero}
\begin{bchapter}
\bauthor{\bsnm{Wang}, \binits{Z.}}:
\bctitle{Zero-shot knowledge distillation from a decision-based black-box model}.
In: \bbtitle{International Conference on Machine Learning},
pp. \bfpage{10675}--\blpage{10685}
(\byear{2021}).
\bcomment{PMLR}
\end{bchapter}
\endbibitem

%%% 42
\bibitem[\protect\citeauthoryear{Min et~al.}{2022}]{min2022pseudo}
\begin{botherref}
\oauthor{\bsnm{Min}, \binits{Z.}},
\oauthor{\bsnm{Ge}, \binits{Q.}},
\oauthor{\bsnm{Tai}, \binits{C.}}:
Why the pseudo label based semi-supervised learning algorithm is effective?
arXiv e-prints,
2211
(2022)
\end{botherref}
\endbibitem

%%% 43
\bibitem[\protect\citeauthoryear{Skeppstedt}{2013}]{skeppstedt2013annotating}
\begin{bchapter}
\bauthor{\bsnm{Skeppstedt}, \binits{M.}}:
\bctitle{Annotating named entities in clinical text by combining pre-annotation and active learning}.
In: \bbtitle{51st Annual Meeting of the Association for Computational Linguistics Proceedings of the Student Research Workshop},
pp. \bfpage{74}--\blpage{80}
(\byear{2013})
\end{bchapter}
\endbibitem

%%% 44
\bibitem[\protect\citeauthoryear{Mikulov{\'a} et~al.}{2023}]{mikulova2023quality}
\begin{botherref}
\oauthor{\bsnm{Mikulov{\'a}}, \binits{M.}},
\oauthor{\bsnm{Straka}, \binits{M.}},
\oauthor{\bsnm{{\v{S}}t{\v{e}}p{\'a}nek}, \binits{J.}},
\oauthor{\bsnm{{\v{S}}t{\v{e}}p{\'a}nkov{\'a}}, \binits{B.}},
\oauthor{\bsnm{Haji{\v{c}}}, \binits{J.}}:
Quality and efficiency of manual annotation: Pre-annotation bias.
arXiv preprint arXiv:2306.09307
(2023)
\end{botherref}
\endbibitem

%%% 45
\bibitem[\protect\citeauthoryear{South et~al.}{2014}]{south2014evaluating}
\begin{barticle}
\bauthor{\bsnm{South}, \binits{B.R.}},
\bauthor{\bsnm{Mowery}, \binits{D.}},
\bauthor{\bsnm{Suo}, \binits{Y.}},
\bauthor{\bsnm{Leng}, \binits{J.}},
\bauthor{\bsnm{Ferr{\'a}ndez}, \binits{O.}},
\bauthor{\bsnm{Meystre}, \binits{S.M.}},
\bauthor{\bsnm{Chapman}, \binits{W.W.}}:
\batitle{Evaluating the effects of machine pre-annotation and an interactive annotation interface on manual de-identification of clinical text}.
\bjtitle{Journal of biomedical informatics}
\bvolume{50},
\bfpage{162}--\blpage{172}
(\byear{2014})
\end{barticle}
\endbibitem

\end{thebibliography}
\end{document}